\documentclass[10pt,twocolumn,letterpaper]{article}

\usepackage[pagenumbers]{cvpr} % To force page numbers, e.g. for an arXiv version

\usepackage{iftex}
\ifXeTeX
  \usepackage{fontspec}
\fi

\usepackage{cuted} %< for strip environment

\usepackage{currfile} %< for \currfilebase macro

\usepackage{caption} %< for "\captionof" commands (teaser)
\usepackage{xspace}
\newcommand{\ours}{Ours\xspace} % no method name (title: "Fill in the Blanks")

\usepackage{booktabs}   % tables generated by scripts/make_tables.py
\usepackage{multirow}
\usepackage{amsmath,amssymb}
\newcommand{\blank}{\ensuremath{\varnothing}} % CTC blank
\usepackage{placeins}   % \FloatBarrier: every table/figure before the conclusion and references
\usepackage{tikz}       % fig/ctc_model_tikz.tex
\usetikzlibrary{positioning,arrows.meta,fit,calc,backgrounds}

\definecolor{cvprblue}{rgb}{0.21,0.49,0.74}
\usepackage[pagebackref,breaklinks,colorlinks,allcolors=cvprblue]{hyperref}

\def\confName{CVPR}
\def\confYear{2027}

\title{Fill in the Blanks: Non-Autoregressive Schema-Parallel\\Key Information Extraction with CTC Decoding}

\author{
Shaked Perek \quad Avihu Dekel \quad Ben Wiesel \quad Sivan Harary \quad Eli Schwartz\\
IBM Research
}

\begin{document}
\maketitle
% Figure 1 (2-column teaser). Left: fig/teaser_left.pdf, rendered from fig/teaser_html/teaser.html. Right: fig/teaser_inset.pdf, from
% scripts/teaser_inset.py (reads the Tab. 1 runs through make_tables.MAIN).
\begin{strip}
\centering
\begin{minipage}[c]{.71\linewidth}\centering
  \includegraphics[width=\linewidth]{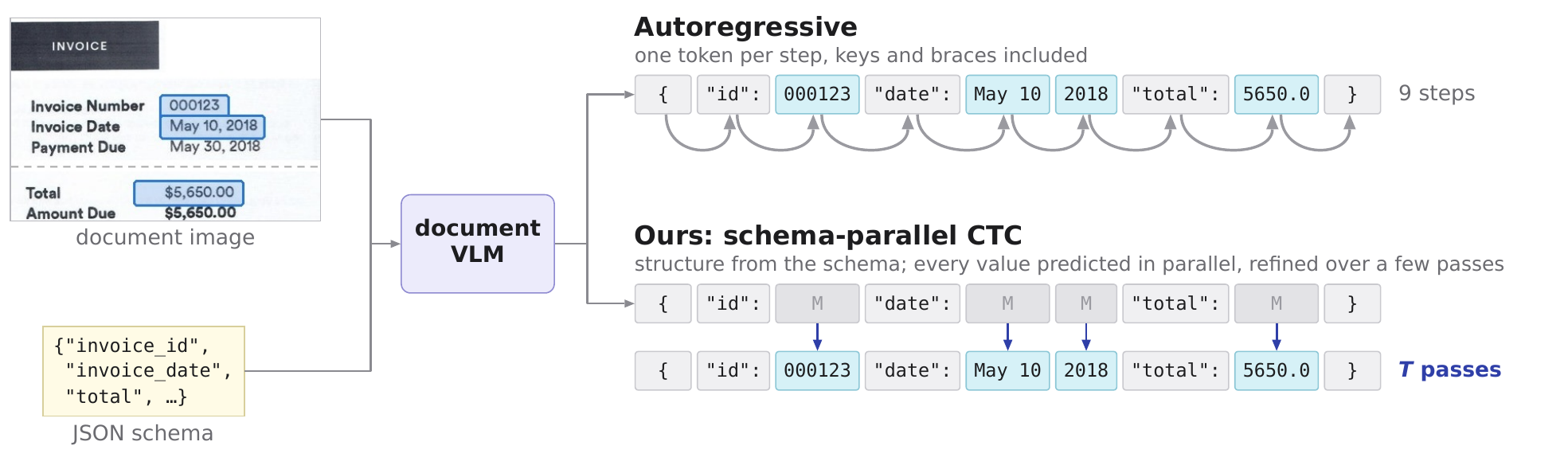}   % from fig/teaser_html/teaser.html (scripts/render_teaser.py)
\end{minipage}\hfill
\begin{minipage}[c]{.27\linewidth}\centering
  \includegraphics[width=\linewidth]{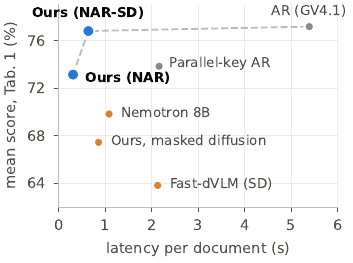}
\end{minipage}
\captionsetup{hypcap=false}
\captionof{figure}{\textbf{All values at once.} A document VLM reads the image and a requested
JSON schema. \emph{Autoregressive} (top): the decoder writes the answer one token per step,
including every key, brace and quote the schema already determines. \emph{Ours, schema-parallel
CTC} (bottom): the structure is written out from the schema, and every value's mask slots are
predicted in parallel and refined over $T$ passes that do not grow with the answer, with CTC
absorbing each value's unknown length. Right: mean Tab.~1 score vs.\
latency.}
\captionsetup{hypcap=true}
\label{fig:\currfilebase}%
\end{strip}

\begin{abstract}
Vision-language models (VLMs) extract structured data from documents by generating the requested
JSON one token at a time, so their sequential steps grow with the length of the answer. Yet in key
information extraction (KIE) the schema already fixes the answer's structure: keys, nesting and
lists are known before decoding, and only the values are unknown. We exploit this with a
non-autoregressive extractor that writes the structure from the schema, sizes every list with a
count pass, and predicts all values in parallel, in a fixed number of passes that does not grow
with the document. Each value is decoded with connectionist temporal classification (CTC) in its
own block of slots, so its length never has to be predicted, and a value that is still open after
its block gets a larger token budget. Zero-shot on VAREX, CORD and FUNSD, our 4B extractor is 3.5$\times$ faster
than the diffusion-VLM competition on average (up to 5.6$\times$) and more accurate, with 4--6
forward passes per document where block diffusion needs 12--74. The same extractor also drafts for
speculative decoding with its own autoregressive base: it runs 8.3$\times$ faster than
autoregressive decoding (up to 10.4$\times$) within 0.4 points of its accuracy, and 3.3$\times$
faster than speculative diffusion decoding while 13.0 points more accurate. We release the model
and code.
\end{abstract}

\section{Introduction}
\label{sec:intro}

% P1 -- the task and why decoding cost matters
Turning a document image into structured data is one of the most common uses of
vision-language models (VLMs). Given an invoice, a form, or a receipt and a JSON schema
describing the fields of interest, the model must return the filled JSON: names, dates,
amounts, addresses, and the rows of every repeated table. This task, key information extraction
(KIE), is where VLMs now outperform pipelines built on separate OCR and layout
models~\cite{layoutlm,layoutlmv3,donut,qwen25vl}. Yet the way VLMs produce the answer is ill-matched to the task. An
autoregressive (AR) decoder writes the JSON one token at a time, including every key, brace, and
quote that the schema had already determined. Its latency therefore grows with the size of the
output, and a dense form with dozens of fields costs dozens of times more decoding steps than a
single answer.

% P2 -- existing parallel decoders and their limits
One line of work reduces this sequential cost. Diffusion VLMs~\cite{you2025lladav,zeng2025diffusionvl,fastdvlm,nvidia2026trimode} replace token-by-token
generation with iterative denoising of masked blocks. In practice they are \emph{block}
diffusion models: the output is split into fixed-size blocks that are denoised one after
another, with bidirectional attention inside a block and causal attention across
blocks~\cite{arriola2025bd3lm}. They still generate the whole JSON, structure included, from left to right,
so their number of sequential steps still grows with the length of the output.

% P3 -- the key observation
We leverage the fact that in KIE the schema already fixes the structure of the answer.
Before decoding begins, the keys, their nesting, and the shape of every list are known. Only
the values are unknown, together with one quantity a parallel decoder cannot see in advance:
their lengths. Values can depend on each other, such as a total on its line items, but most of
what a field needs is the document and the schema. We therefore predict every value at once and
let the values inform each other over a few refinement passes (\cref{fig:teaser}).

% P4 -- the method
We build a non-autoregressive extractor that works from the schema alone. Two things are
unknown before decoding: how long each value is, and how many items each list has. We address both
in one decoding scheme. The schema is written out as a skeleton in which every value is a block of
mask slots, sized by a count pass that predicts the number of items in each list, and a
bidirectional extractor predicts all slots in parallel, refining them over a few confidence passes. Each block is decoded with
connectionist temporal classification (CTC)~\cite{graves2006ctc}: a value may occupy any number of
slots up to the block size, with blank symbols absorbing the rest, so no value length is predicted
or given. A value still open after its block gets a larger token budget: its decoded text is fixed, a
fresh block is appended, and the next forward continues the value (a \emph{continuation}). No gold keys, value lengths or item counts are needed at inference. The same
extractor also works as a speculative drafter for its own autoregressive base
model~\cite{leviathan2023,chen2023}. Because the schema fixes where every value sits, the base
model verifies the drafted values of all keys in one forward per round, rather than one block at a
time from left to right, which trades part of the speedup for autoregressive-level accuracy.

% P5 -- results, as highlights (numbers live in Sec. 4)
We evaluate on VAREX, CORD and FUNSD, KIE benchmarks that the model never saw during training, under a single protocol in which every method receives only the image and the
benchmark's schema, and with all methods timed on the same GPU. Our extractor, fine-tuned from Granite Vision 4.1 (4B)~\cite{granite41vision}, is more accurate
than an 8B block-diffusion VLM on all three benchmarks, and 3.5$\times$ faster on average (up to
5.6$\times$). It is also substantially faster than parallel-key autoregressive extraction with the
same base model. In its speculative mode it approaches the accuracy of autoregressive decoding
(\cref{tab:main}).

% Contributions
In summary, our contributions are:
\begin{itemize}
  \item Schema-parallel decoding: once the schema is known no generation order is needed, so
        every value is predicted at once and the number of sequential steps stays nearly
        constant per document instead of growing with the output.
  \item Variable-length values from a schema-only prompt: CTC value slots, a token budget
        that grows for values still open, and a list-count pass, with no gold keys, lengths, or item
        counts at inference.
  \item A comparison on three zero-shot benchmarks against autoregressive, parallel-key
        autoregressive, and diffusion VLMs, under one schema-only protocol and on the same
        hardware.
  \item A non-autoregressive KIE extractor built on Granite Vision 4.1, released with its
        weights and code; to our knowledge the first released non-autoregressive model for KIE.
\end{itemize}

\section{Related Work}
\label{sec:related}

\paragraph{Key information extraction.}
Early KIE systems on document images run OCR first and then classify or link the recognized
tokens, using layout-aware encoders such as the LayoutLM family~\cite{layoutlm,layoutlmv3}.
OCR-free models instead read the image directly and generate the answer as a sequence, either from
task-specific pretraining (Donut~\cite{donut}, Pix2Struct~\cite{pix2struct}) or as one capability
of a general document VLM~\cite{qwen25vl,internvl,granite41vision}. We build on a document VLM and
follow the schema-driven formulation: the model receives the image and a JSON schema and returns
the filled JSON. Unlike these models, we decode that JSON without generating it left to right.

\paragraph{Diffusion and block-diffusion language models.}
Masked diffusion language models generate text by iteratively unmasking a fully masked
sequence~\cite{sahoo2024mdlm,nie2025llada}. Block diffusion~\cite{arriola2025bd3lm} splits the
output into fixed-size blocks that are denoised one at a time, bidirectional within a block and
causal across blocks, which restores KV caching and variable-length generation. This design has
been carried over to VLMs, either trained as diffusion models from the start~\cite{you2025lladav}
or converted from autoregressive VLMs (DiffusionVL~\cite{zeng2025diffusionvl},
Fast-dVLM~\cite{fastdvlm}), and combined with autoregressive and self-speculative decoding in a
single model~\cite{nvidia2026trimode}. Our extractor shares their mask tokens and bidirectional
attention but differs in two ways. First, these models still generate the whole answer,
structure included, and visit the blocks left to right, so their sequential steps grow with the
length of the output; we write the structure from the schema and decode every value in the same
passes, in no order. Second, their output length is decided by where the generation ends, not
per value; we decode each value with CTC, which leaves its length to the model within its own
slots.

\paragraph{Structured generation.}
The usual way to make a language model emit valid JSON is constrained decoding: at every
autoregressive step, tokens that would leave a grammar or JSON schema are masked
out~\cite{geng2023gcd,willard2023outlines,dong2025xgrammar}. This guarantees well-formed output, and jump-forward decoding~\cite{zheng2023sglang} skips the
structure tokens that the grammar forces, but the values are still generated one token at a time,
one after another. Parallel-key autoregressive decoding, our strongest autoregressive baseline
(\cref{sec:exp_setup}), skips the structure and decodes the values in parallel (lists still as JSON), so its latency is a
lower bound for any template-forced autoregressive decoder. When the structure is known before
decoding, it need not be generated at all. S$^3$~\cite{xiong2025s3} takes this view for diffusion
LLMs: without any training, it writes the schema template into the output of a text-only masked
diffusion model, which fills the values, with a \texttt{null} token for absent fields. It is
evaluated on a seven-field text extraction task (WikiBio), and its authors note that the results
depend on how many mask tokens each field receives. We share the schema template, but train a
document VLM to fill it from the image, decode each value with CTC so that its length does not
depend on the number of slots, extend the token budget of values still open, and size every list
from the image. A different way to use the structure keeps the autoregressive decoder and runs one
short answer per field, all in parallel~\cite{lu2024padellm}. A shared prompt prefix makes this
cheap~\cite{juravsky2024hydragen,zheng2023sglang}, and we use it, on our base model, as our
strongest autoregressive baseline.

\paragraph{Parallel decoding for KIE.}
Closest to our setting, PIP~\cite{wang2026pip} introduced parallel masked-value prediction for
KIE: it replaces the values in the prompt with mask tokens, switches the VLM to bidirectional
attention, and predicts every masked position in one forward pass, each position decoded
independently. Its fixed-slot formulation assigns one token per slot, which leaves open how to
decode values of unknown length, values longer than the slots, and lists whose number of items is
not known in advance; these are what schema-only extraction requires. No model or code for PIP
is available, so we compare against a matched fixed-slot decoder, trained on our base model and
data (\cref{sec:fixed_slot}).

\paragraph{Non-autoregressive generation with CTC.}
Non-autoregressive machine translation predicts the target in parallel and must handle the unknown
target length, either by predicting it explicitly~\cite{gu2018nat} or by decoding a longer
canvas with connectionist temporal classification (CTC)~\cite{graves2006ctc}, which
marginalizes over alignments and absorbs length with blank symbols~\cite{libovicky2018,saharia2020,chan2020imputer}; Mask-CTC adds iterative refinement of low-confidence tokens on top of a CTC
draft~\cite{higuchi2020maskctc}. Non-autoregressive models have also been applied to
vision-language tasks such as captioning and VQA~\cite{shi2024narvl,kim2023magvlt}, and parallel
decoding is common in scene-text recognition~\cite{bautista2022parseq}. We apply CTC per value slot rather than over a
whole sequence: the schema fixes where each value sits, CTC handles each value's length, and
continuations extend values past the token budget of one block.

\paragraph{Speculative decoding.}
Speculative decoding drafts several tokens with a cheap model and verifies them with the target
model in one forward pass, preserving the target's output~\cite{leviathan2023,chen2023},
and has been extended to VLMs~\cite{specvlm}; for document OCR, a single model can draft with block diffusion and verify autoregressively~\cite{gravityocr}. Self-speculative diffusion VLMs~\cite{fastdvlm,nvidia2026trimode} draft and verify one block
at a time, left to right. Our second mode drafts with the non-autoregressive extractor for its own autoregressive base
and, since the schema fixes where every value sits, verifies the drafted values of all keys in
one forward.

\section{Method}
% Method overview (2-column). Drawing: fig/ctc_model_tikz.tex; preview: scripts/preview_fig.tex.
\begin{figure*}[t]
\centering
\resizebox{\linewidth}{!}{% CTC model overview (tikzpicture only; wrapped by fig/ctc_model.tex, previewed by
% scripts/preview_fig.tex). Needs amssymb (\varnothing), tikz + libraries positioning,
% arrows.meta, fit, calc, backgrounds. Example values are illustrative; each value block is
% drawn with 5 of its 20 slots. Four panels (a)-(d) with explicit edges and 6 mm gaps, so the
% arrows between them are visible; ~20.6 cm wide, scaled ~0.85 to \textwidth.
\begin{tikzpicture}[
  font=\sffamily\scriptsize, >={Stealth[length=5pt,width=4pt]},
  slot/.style={draw=black!40, minimum width=6.6mm, minimum height=4.6mm, inner sep=0pt,
               font=\sffamily\scriptsize},
  msk/.style={slot, fill=black!8, text=black!45},
  tok/.style={slot, fill=cyan!15},
  blk/.style={slot, fill=white, text=black!40},
  new/.style={slot, fill=orange!25},
  key/.style={font=\ttfamily\scriptsize, anchor=east},
  res/.style={font=\ttfamily\scriptsize, anchor=west},
  panel/.style={draw=black!35, rounded corners=2pt, fill=black!1},
  title/.style={font=\sffamily\footnotesize\bfseries, anchor=north west, inner sep=3pt},
  note/.style={font=\sffamily\scriptsize, text=black!60},
  flow/.style={->, very thick, black!60},
]
\def\w{0.70}    % slot pitch
\def\top{0.75}  % panel top edge
\def\bot{-4.4}  % panel bottom edge
\def\ym{-1.45}  % height of the arrows between panels
% panel x-extents: (a) 0-3.7, (b) 4.3-10.45, (c) 11.05-15.0, (d) 15.6-22.0
\begin{scope}[on background layer]
  \node[panel, fit={(0,\top) (3.7,\bot)}, inner sep=0pt] (A) {};
  \node[panel, fit={(4.3,\top) (10.45,\bot)}, inner sep=0pt] (B) {};
  \node[panel, fit={(11.05,\top) (15.0,\bot)}, inner sep=0pt, fill=blue!6, draw=blue!50!black!45] (C) {};
  \node[panel, fit={(15.6,\top) (22.0,\bot)}, inner sep=0pt] (D) {};
\end{scope}
\node[title] at (A.north west) {(a) inputs, count pass};
\node[title] at (B.north west) {(b) skeleton: $n$ \texttt{[M]} slots per value};
\node[title] at (C.north west) {(c) bidirectional VLM, $\times T$};
\node[title] at (D.north west) {(d) parallel CTC decoding};
\foreach \p/\q in {A/B, B/C, C/D} \draw[flow] (\p.east |- 0,\ym) -- (\q.west |- 0,\ym);

% ======================= (a) inputs and the count pass =======================
\node[draw=black!45, fill=black!3, minimum width=8mm, minimum height=12mm] (doc) at (0.5,-0.4) {};
\foreach \y/\l in {0.4/0.5,0.25/0.35,0.1/0.55,-0.1/0.55,-0.25/0.55,-0.4/0.3}
  \draw[black!30, line width=1pt] ($(doc.center)+(-0.28,\y)$) -- ++(\l,0);
\node[note, below=0.4mm of doc, align=center] {doc.\\image};
\node[draw=black!45, fill=yellow!10, align=left, font=\ttfamily\scriptsize, inner sep=1.5pt,
      anchor=north west] (schema) at (1.0,0.2)
  {\{"vendor": str,\\\ "date": str,\\\ "lines": [\{\\\ \ "desc": str,\\\ \ "qty": num\}]\}};
\node[note, below=0.4mm of schema] {JSON schema};
\node[draw=green!45!black!55, rounded corners=2pt, fill=green!8, align=center, inner sep=3pt,
      text width=27mm] (count) at (1.85,-3.4)
  {\textbf{count pass}\\[1pt] how many items in each list?\\[1pt]\texttt{lines}: 2 items};
\draw[->, thick, black!55] (doc.south |- 0,-2.0) -- (doc.south |- count.north);
\draw[->, thick, black!55] (schema.south |- 0,-2.0) -- (schema.south |- count.north);

% ======================= (b) skeleton =======================
\def\xs{6.75}
\foreach \r/\k in {0/vendor,1/date,2/lines[0].desc,3/lines[0].qty,4/lines[1].desc,5/lines[1].qty} {
  \node[key] at (\xs, -0.3-0.62*\r) {"\k":};
  \foreach \c in {0,...,4} \node[msk] (m\r\c) at (\xs+0.45+\w*\c, -0.3-0.62*\r) {M};
}
\node[note, align=center] at (7.375, -4.05)
  {one block for every value of both \texttt{lines} items};

% ======================= (c) the model: T confidence passes =======================
% One value block (the "vendor" row of (b)), labelled by its state AFTER pass t (t=0 = input).
% Passes 1..T-1 reveal slots whose argmax is non-blank with p > tau (evaluate.py run_nar_batch);
% blanks are never revealed. Pass T is a plain argmax over every slot, then CTC collapse (d).
% Outlined = filled by that pass. Every pass is one forward over all values.
\def\cw{0.62}                 % slot pitch in (c)
\def\cx{12.15}\def\gyc{0.02}  % first slot
\def\rp{0.56}                 % row pitch
\tikzset{cs/.style={slot, minimum width=6.1mm, minimum height=4.2mm, font=\sffamily\tiny},
         cmk/.style={cs, fill=black!8, text=black!45},
         ct/.style={cs, fill=cyan!15}, cb/.style={cs, fill=white, text=black!40},
         fresh/.style={draw=cyan!55!black, very thick}, ctf/.style={ct, fresh},
         fin/.style={draw=blue!55!black, thick, densely dashed}, ctF/.style={ct, fin}, cbF/.style={cb, fin}}
\def\crow#1#2{\foreach \t/\s [count=\c from 0] in {#2} \node[\s] (r#1\c) at (\cx+\cw*\c, -0.38-\rp*#1) {\t};}
\crow{0}{M/cmk,M/cmk,M/cmk,M/cmk,M/cmk}
\crow{1}{Acme/ctf,M/cmk,M/cmk,M/cmk,M/cmk}
\crow{2}{Acme/ct,Acme/ctf,M/cmk,Corp/ctf,M/cmk}
\crow{3}{Acme/ct,Acme/ct,$\varnothing$/cbF,Corp/ct,$\varnothing$/cbF}
\foreach \r in {0,...,3} \node[note, anchor=east] at ($(r\r0.west)+(-0.02,0)$) {$t{=}\r$};
\node[note, anchor=west] at (11.15,\gyc) {\texttt{"vendor"} after each pass};
\node[note, align=left, anchor=north west] at (11.15,-2.3)
  {each pass: \textbf{one forward}\\over all values at once};
% legend: swatch and text on the same baseline (explicit rows, not inline \tikz)
\tikzset{lgR/.style={fill=cyan!15, draw=cyan!55!black, very thick},
         lgM/.style={fill=black!8, draw=black!40},
         lgB/.style={fill=white, draw=blue!55!black, thick, densely dashed}}
\foreach \y/\sty/\txt in {-3.25/lgR/{revealed ($p{>}\tau$), fixed},
                          -3.62/lgM/{masked, re-predicted},
                          -3.99/lgB/{blank}} {
  \draw[\sty] (11.3,\y-0.12) rectangle (11.62,\y+0.12);
  \node[note, anchor=west] at (11.68,\y) {\txt};
}

% ======================= (d) CTC decoding =======================
\def\xp{16.15}
\def\row#1#2#3{\foreach \t [count=\c from 0] in {#3} \node[tok] (p#1\c) at (\xp+\w*\c, #2) {\t};}
\row{0}{-0.3}{Acme,Acme,,Corp,}
\row{1}{-0.92}{03/11,/,2024,,}
\row{2}{-1.54}{Steel,bolts,M8,x,40}
\row{3}{-2.16}{12,,,,}
\foreach \n in {p02,p14,p31,p32,p34} \node[blk] at (\n) {$\varnothing$};
\node[tok] at (p04) {\texttt{"}};
\node[tok] at (p13) {\texttt{"}};
\node[tok] at (p33) {\texttt{"}};
\def\xc{\xp+\w*4+0.6}
\node[res] (c0) at (\xc,-0.3) {Acme Corp};
\node[res] (c1) at (\xc,-0.92) {03/11/2024};
\node[res, text=orange!75!black, font=\sffamily\scriptsize] (c2) at (\xc,-1.54) {full: go on};
\node[res] (c3) at (\xc,-2.16) {12};
\foreach \r in {0,...,3} \draw[->, black!50] ($(p\r4.east)+(0.04,0)$) -- ($(c\r.west)+(-0.02,0)$);
\node[note, anchor=south west] at ($(p00.north west)+(0,0.4mm)$)
  {CTC collapse: merge repeats, then drop $\varnothing$};
% continuation window: the draft is committed as ordinary tokens, then fresh slots
\node[res, text=black!55, inner xsep=0pt] (qold) at (15.8,-2.85) {Steel bolts M8 x 40};
\foreach \t [count=\c from 0] in {mm,{,},zinc,}
  \node[new] (q\c) at ($(qold.east)+(0.42+\w*\c,0)$) {\t};
\node[new] at (q3) {\texttt{"}};
\draw[->, orange!75!black, thick, rounded corners=2pt]
  (c2.east) -- ++(0.15,0) |- ($(q3.north)+(0,0.2)$) -- (q3.north);
\node[note, anchor=north west, align=left] at ($(qold.south west)+(0,-0.6mm)$)
  {\textbf{budget $+n$}: fix the draft,\\append fresh slots, re-run $\rightarrow$\\
   \texttt{Steel bolts M8 x 40 mm, zinc}};
\end{tikzpicture}}
\caption{\textbf{Schema-parallel CTC decoding.} (a) A count pass reads the image and the schema
and sizes every list. (b) The schema is written out as a skeleton in which every value gets a
block of $n$ \texttt{[M]} slots. (c) A bidirectional forward predicts all slots of all
values at once; this is repeated for $T$ passes, each but the last revealing the slots
predicted with probability above $\tau$ and re-predicting the rest (blanks are never revealed; the last pass takes the argmax of every
slot), with the
image--prompt key/values computed once. (d) CTC collapse first merges repeats, then drops blanks
($\varnothing$); a value still open after its block gets $n$ more slots (budget $+n$).}
\label{fig:\currfilebase}
\end{figure*}
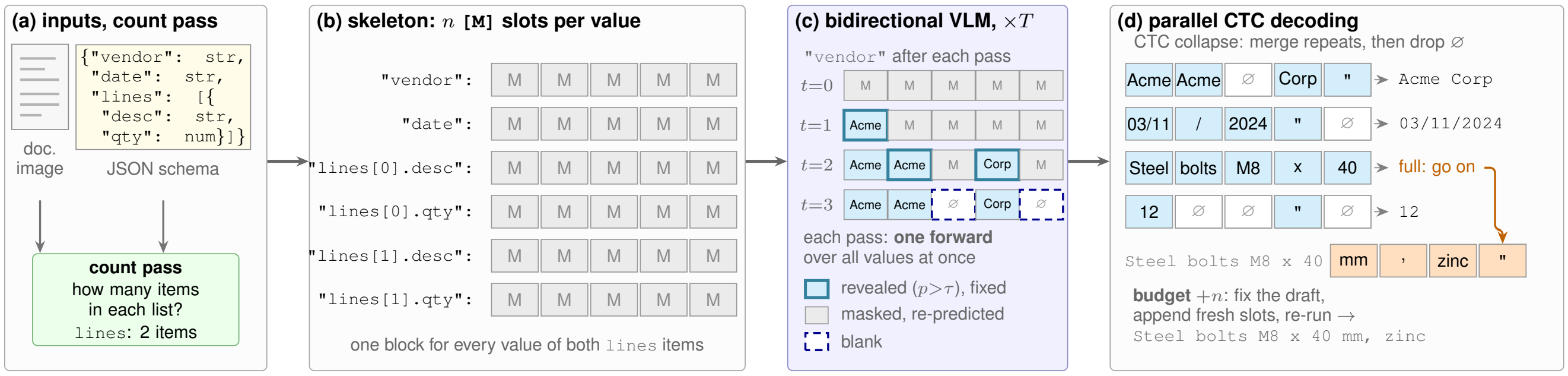

\label{sec:method}

\subsection{Problem setup}
\label{sec:setup}
A KIE query is a document image $I$ and a JSON schema $S$. The schema lists the keys, their
types, the nesting of objects and which fields are lists of objects; the answer is the JSON
object $y$ that fills every leaf of $S$ with its value from $I$, or a null value when the field
is absent. We follow a schema-only protocol at test time: the model receives $I$ and $S$ and
nothing else --- no gold keys, no value lengths and no number of items in each list. An
autoregressive VLM produces $y$ token by token, structure included. Our observation is that
everything in $y$ except the values, and the number of list items, is already determined by $S$.
We therefore build the structure from $S$ and predict only the values, all of them in
parallel (\cref{fig:ctc_model}).

\subsection{Schema skeleton and list counts}
\label{sec:skeleton}
Given item counts for every list in $S$, we serialize the answer JSON with each leaf value
replaced by a block of $n$ mask tokens. Keys, braces, commas and quotes are written out as fixed
tokens (read but not predicted), so the model never has to generate structure. The item counts are the one
part of the structure the schema does not give. We predict them with a small count adapter, a
LoRA on the extractor's weights that is switched on only for this pass: a count query lists the
schema path of every list, each followed by a single mask slot, and one bidirectional forward
predicts all counts at once. Nested lists need one count forward per
nesting level, since an inner list exists once per predicted outer item. The skeleton is then
expanded with the predicted items. Counts come before values, so the count pass adds one sequential step per level of list nesting;
like every other step of the method (the confidence passes and the budget extensions), it does not
depend on the length of the output or the number of keys. The adapter is trained for one epoch on count queries built from the same
training data.

\subsection{CTC value slots}
\label{sec:ctc}
The extractor is a pretrained document VLM whose causal attention is replaced by
bidirectional attention, so every position, including every mask slot, attends to the whole
image, prompt and skeleton. Each of a value's $n$ slots predicts one symbol: a token of the
vocabulary, or a blank \blank\ meaning that the slot emits nothing. A value
$v=(v_1,\dots,v_m)$ has its own length $m\le n$, which differs from value to value, so the slots
need not match the tokens one to one. Following CTC~\cite{graves2006ctc}, an assignment
$a=(a_1,\dots,a_n)$ of symbols to the slots, called an alignment, is read out by the collapse map
$\mathcal{B}$, which merges repeated tokens and then removes blanks. The block is trained to maximize the total probability of all alignments that
collapse to $v$; the CTC loss is then:
\begin{equation}
  \mathcal{L}_{\text{CTC}}(v) = -\log \sum_{a\,:\,\mathcal{B}(a)=v}\ \prod_{i=1}^{n} p(a_i \mid I, S, \tilde{y}),
\end{equation}
where $\tilde{y}$ is the partially masked skeleton. Within one forward the slot symbols are
conditionally independent given $(I, S, \tilde{y})$; dependencies between slots and between values
enter through the confidence reveals (below), which change $\tilde{y}$ between forwards. The loss is summed over all values of the document. Because the loss sums over
all alignments, one block of $n$ slots holds a value of any length up to $n$: unused slots emit
blanks, so the length need not be known before decoding, as it must be when each token has its own slot. We reuse the mask token as the blank, so an unused slot is simply a slot that stays
masked. Absent fields are trained to the literal value \texttt{NA}. At inference, each block is
decoded by taking the argmax at every slot and applying $\mathcal{B}$.

\paragraph{Confidence reveals.}
To let slots and values condition on each other, we run $T$ forwards. After each of the first
$T{-}1$, slots whose argmax is not blank and whose probability exceeds $\tau$ are written into the
input as fixed tokens, and the next forward predicts the remaining slots with that context; the
last forward takes the argmax of every slot, so $T{=}1$ is a single pass with no reveals. Training mirrors inference: we run the
same $T$ forwards with the model's own reveals and apply the CTC loss at every forward.
Unlike an autoregressive decoder, the reveals impose no order; any value, and any slot of a
value, can be committed in any pass.

\subsection{Token budget and continuation}
\label{sec:continuation}
A block of $n$ slots bounds the length that one decode can emit.
To handle longer values without predicting their lengths in advance,
we progressively extend each value's token budget by $n$ slots.
As illustrated in Fig.~\ref{fig:ctc_model}(d), a continuation
writes the collapsed draft $\mathcal{B}(a)$ into the input as
fixed tokens, followed by a fresh block of $n$ mask slots.
The next forward predicts the continuation of all values
in parallel.

\paragraph{Training.}
To teach the model to continue from arbitrary positions, we do
not always start with an empty value. For each value independently,
we uniformly sample a ground-truth prefix length, reveal that
prefix, and append $n$ mask slots. The CTC loss supervises
the remaining suffix. We then decode the model's predictions
and use its own draft, including mistakes, as the prefix for
a continuation. The new slots are supervised with the remaining
ground-truth tokens after the emitted draft.
Every value is extended once during training; values that are
already complete receive an all-blank target, teaching the model
to emit nothing once a value is finished. For values longer than
the available block ($m > n$), the CTC target is truncated to
the first $n-1$ remaining tokens, with the rest assigned to
the continuation.

\paragraph{Inference.}
We initialize every value with $n$ mask slots and decode using
the confidence passes of Sec.~3.3 (Fig.~\ref{fig:ctc_model}(c)).
We then extend every value by appending $n$ fresh slots after
its collapsed draft, including values that are already complete.
After this extension, only values whose text remains open,
indicated by the absence of a closing quote (\texttt{"}) as
illustrated in Fig.~\ref{fig:ctc_model}(d), receive another
block of $n$ slots, up to a maximum budget of $3n$.
At inference, we collapse the entire block, including the
committed draft and the new slots, allowing the model to revise
earlier predictions and repair errors from the confidence
passes. All open values are extended in the same forward,
so continuation does not introduce a sequential dependency
between fields.

\subsection{Parallel Decoding and Dependencies}
\label{sec:parallel}
Every value is decoded in the same forwards, so the number of forwards per document is fixed
by the decoding schedule ($T$ confidence passes, up to two continuations, plus one count
forward per list level), not by the number of keys or the length of the answer (\cref{sec:steps}).
Values are not assumed independent, since all slots attend to each other and the reveals pass
fixed tokens between values; we drop only a fixed generation order, and where order matters the
speculative mode (\cref{sec:sd}) verifies left to right.

\subsection{Efficient inference}
\label{sec:inference}
The image and prompt make up most of the sequence, but only the JSON changes between passes.
We run the first value forward with full bidirectional attention over image, prompt and
skeleton, so the image representation is computed while seeing the masked skeleton, and then
freeze the key/value (KV) cache of the image and prompt. Every later forward (the remaining
confidence passes and the continuations) runs only over the JSON
tokens, attending bidirectionally to the frozen KV cache and to each other. 
% Freezing is valid
% because the image and prompt do not change between passes, and every later pass still reads the
% image representation the first pass computed while attending to the skeleton; it removes the image
% and prompt from every later forward. 
This stale prefix keeps
the accuracy of full bidirectional decoding at every step, whereas training with a prefix-causal mask, so that the image never attends to the
skeleton, loses 2.2 and 3.4 points on VAREX and CORD (\cref{tab:design}a).

\subsection{Speculative decoding mode}
\label{sec:sd}
The same extractor can draft for its own autoregressive base model. In each \emph{verification
round}, the base model
verifies the drafted values of all keys in one forward over the JSON, with the image and
prompt cached. 
% Because the schema fixes where every value sits in the JSON, one verification
% forward checks every value of the document at once, unlike block-wise self-speculative
% decoding, which verifies one block at a time from left to right. 
For each value it accepts the longest prefix on which the draft matches its own
greedy prediction, plus its next token as a correction; a value is finished when that correction
closes the string. The extractor then redrafts the remaining keys from their accepted prefixes,
again in one forward for all of them. Verification makes each value agree with the base model's
greedy prediction for that value, given the other values' drafts in its context rather than the
base model's own earlier output, and up to tokenization differences, which can stall a value that the draft had right;
for example, a draft that writes \texttt{2024} as one token is rejected when the base model would
write it as \texttt{20}+\texttt{24}, although the text is the same. Values still open when the cap on verification rounds is reached therefore fall back to their first
non-autoregressive draft.

\section{Experiments and Results}
\label{sec:experiments}

% GENERATED by scripts/make_tables.py -- do not edit by hand.
\begin{table}[t]
\centering
\setlength{\tabcolsep}{3pt}
\resizebox{\linewidth}{!}{%
\begin{tabular}{lrrrrrr}
\toprule
 & \multicolumn{2}{c}{VAREX} & \multicolumn{2}{c}{CORD} & \multicolumn{2}{c}{FUNSD} \\
Method & EM & ms & F1 & ms & F1 & ms \\
\midrule
GV4.1 AR & 92.1 & 6,113 & 80.3 & 3,986 & 59.1 & 6,067 \\
\midrule
Parallel-key AR & 91.0 & 1,320 & 75.4 & 3,298 & 55.1 & 1,859 \\
Nemotron 8B & 85.8 & 945 & 71.1 & 902 & 52.6 & 1,399 \\
\ours, masked diff. & 79.9 & 793 & 69.6 & 779 & 52.8 & 1,000 \\
\ours NAR & \textbf{88.4} & \textbf{278} & \textbf{76.8} & \textbf{397} & \textbf{54.3} & \textbf{249} \\
\midrule
Fast-dVLM 3B & 72.4 & 2,500 & 71.1 & 1,695 & 48.0 & 2,188 \\
\ours NAR-SD & \textbf{91.0} & \textbf{646} & \textbf{80.4} & \textbf{686} & \textbf{59.0} & \textbf{582} \\
\bottomrule
\end{tabular}}
\caption{\textbf{Accuracy and latency on three zero-shot benchmarks.} Every method gets only the image and the benchmark's schema. VAREX: VAREX-Bench score (official scorer); CORD: field F1; FUNSD: pair F1. Latency: mean ms per document over all documents, one document at a time (\cref{sec:exp_setup}). NAR: token budget up to 3$n$ per value; NAR-SD: a two-pass draft and up to three verification rounds, values still open at the cap fall back to their first draft. Averages in the text are geometric means of the per-benchmark speedups. \emph{Ours, masked diff.} is our model with its CTC slots replaced by masked diffusion with 8 steps (\cref{tab:decoder}).}
\label{tab:main}
\end{table}
     % generated by scripts/make_tables.py (declared early: floats)
\subsection{Setup}
\label{sec:exp_setup}
\paragraph{Training data.}
We train on a synthetic dataset built from real documents. Pages are selected from the English
subset of FinePDFs~\cite{finepdfs}: mostly forms and transactional documents such as invoices,
delivery notes and statements, but also product sheets, technical drawings and listings. For each
page, an autoregressive VLM, given the page image and the text extracted programmatically from the
PDF, proposes a schema and reads its values, so every document comes with its own schema. Each
document is paired with several instructions that vary how the schema is presented (as a JSON
Schema, a typed model, a table or a list of field names), which fields are requested and in what
order, and the null and formatting conventions; the training target is always the answer JSON,
with its values masked. We train on 150K documents and keep a held-out split for validation, on
which the decoding settings were selected before any public benchmark was evaluated.

\paragraph{Zero-shot evaluation.}
We evaluate on VAREX~\cite{barzelay2026varex} (1,777 documents with schema-driven, nested
extraction), CORD~\cite{cord} (100 receipts with line items) and FUNSD~\cite{funsd} (key--value pairs, with the revised labels of~\cite{vu2020funsdrev}), public benchmarks none of which is used for training or model
selection. Every
method receives only the document image and the benchmark's own JSON schema in the same
prompt; no method sees gold keys, value lengths or list sizes. CORD and FUNSD are
converted to the same schema-and-answer format as VAREX. We report the VAREX-Bench score on
VAREX, computed with the benchmark's official scorer, and the metrics used in the literature for the
others: field F1 on CORD and pair F1 on FUNSD.

\paragraph{Implementation.}
All models in the paper start from Granite Vision (GV) 4.1 4B~\cite{granite41vision}. GV 4.1 was already trained on a larger KIE dataset encompassing all 150K documents used in our fine-tuning. Thus, our NAR model gains no additional data exposure over the AR baseline.. The extractor is trained with LoRA (rank 16, $\alpha{=}32$) for five epochs on one node with eight H200 GPUs
(AdamW, learning rate $2{\cdot}10^{-5}$ with a cosine schedule and 5\% warmup, global batch 32,
bf16), bidirectional
attention, $n{=}20$ CTC slots per value and $T{=}3$ confidence passes with reveal threshold $\tau{=}0.7$; the count adapter is a
separate rank-16 LoRA, trained for one epoch on count queries only, on top of the frozen extractor,
and switched on only for the count pass. At inference the extractor uses the stale prefix and a
token budget of up to $3n$ per value (\cref{sec:continuation}), and the speculative mode drafts with two
passes ($T{=}2$) and runs up to three verification rounds. Latency is the mean time per document over all documents of a
benchmark, one document at a time on a single H200 GPU, every method on the same exclusive
node.

\paragraph{Baselines.}
We compare with two autoregressive and two diffusion decoders, each run with its released
inference code and the same prompt. \emph{AR} is Granite Vision 4.1 itself, generating the
whole JSON with greedy decoding; it is the accuracy reference and the model our extractor is
fine-tuned from. \emph{Parallel-key AR} keeps the same model but removes the structure from the
generation: the image and prompt are prefilled once, and one short greedy answer is decoded for
every key of the schema~\cite{lu2024padellm}, all keys as one batch from the shared
prefix~\cite{juravsky2024hydragen,zheng2023sglang}; each
list becomes one query that returns the whole list as JSON. It is the strongest autoregressive
baseline for latency. \emph{Nemotron diffusion 8B}~\cite{nvidia2026trimode} is a block-diffusion
VLM that decodes blocks of 32 tokens in parallel, committing every token above confidence 0.9
at each step, with no verifier; it is the closest non-autoregressive competitor of our
extractor. \emph{Fast-dVLM 3B}~\cite{fastdvlm}, converted from Qwen2.5-VL~\cite{qwen25vl},
drafts each block with diffusion and verifies it with its causal mode, so we compare it with our
speculative mode. As a controlled comparison, we also report our own model with its CTC slots
replaced by masked diffusion (8 steps), trained on the same base, data and skeleton (\cref{sec:fixed_slot}).

\subsection{Main results}
\label{sec:main_results}
\Cref{tab:main} compares each of our two modes with the methods of the same kind.

\paragraph{Without a verifier.}
Our extractor is more accurate than Nemotron diffusion 8B on all three benchmarks (+3.3 points on
average) and 3.5$\times$ faster on average (up to 5.6$\times$). The gap is not model size: Nemotron's
forwards are cheaper than ours (22--26 ms against 55--76 ms), but a median document needs 12--74 of
them against our 4--5 (\cref{sec:steps}). With the same base model and data, replacing our CTC
slots with masked diffusion (8 steps) is 2.8$\times$ slower and 5.7 points less accurate on average.
Parallel-key AR on our base model is 6.6$\times$ slower on average and 0.7 points more accurate.

\begin{figure*}[!t]
\centering
\begin{minipage}[c]{.36\linewidth}\centering
{\setlength{\fboxsep}{0pt}\setlength{\fboxrule}{0.4pt}\fbox{\includegraphics[width=\dimexpr\linewidth-0.8pt\relax]{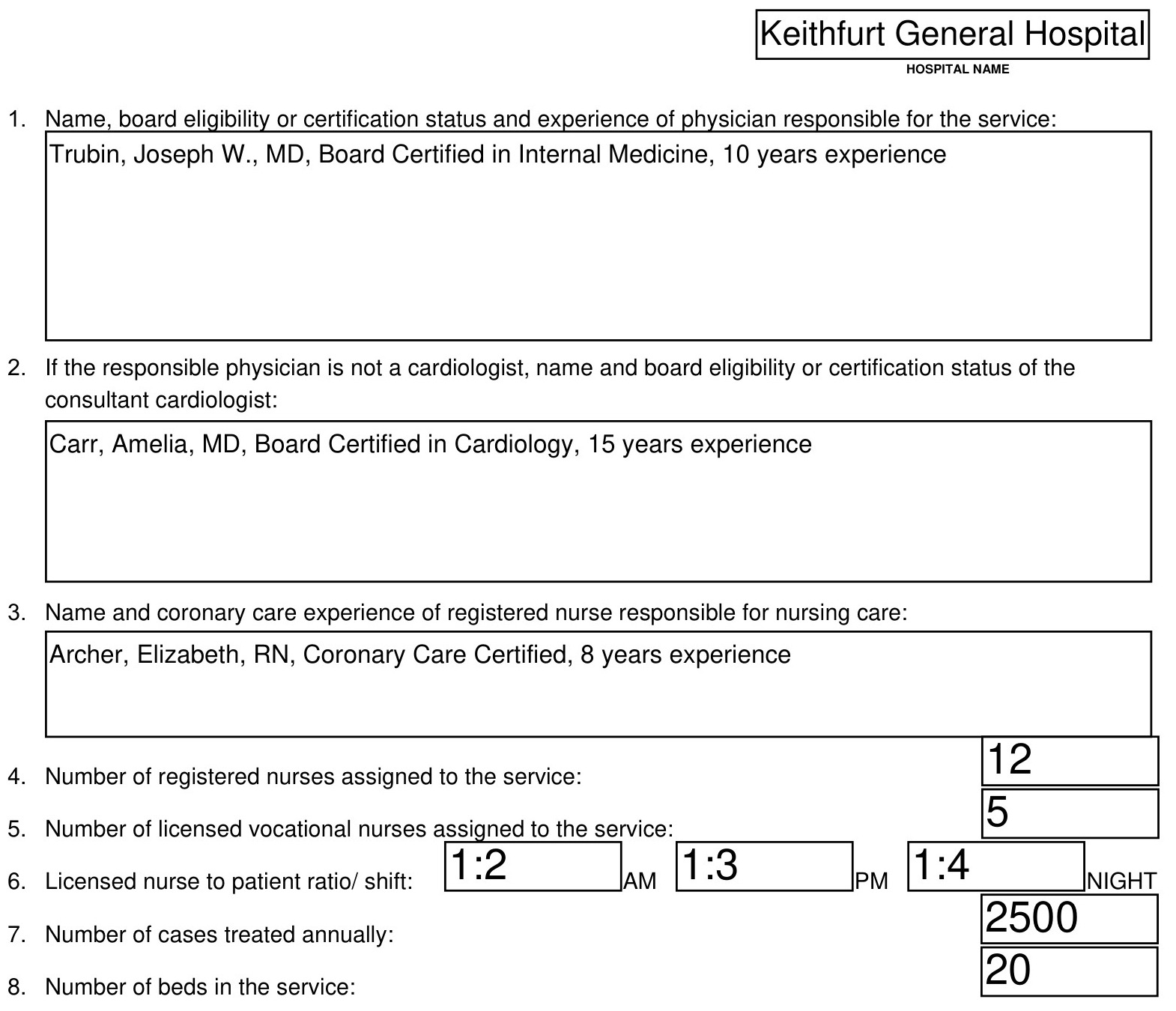}}}\\[2pt]
{\scriptsize Input: a filled CDPH application form (crop)}
\end{minipage}\hfill
\begin{minipage}[c]{.62\linewidth}\centering
\includegraphics[width=\linewidth]{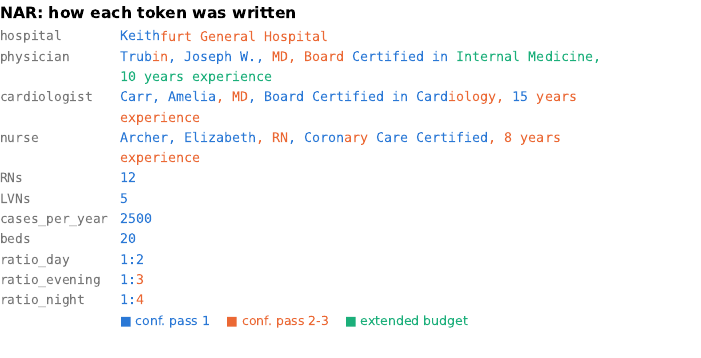}
\end{minipage}
\caption{\textbf{Decoding trace} on a VAREX test form, coloured by the forward that wrote each
token. The first bidirectional pass commits most of every value (blue) and the later confidence
passes fill what it was unsure of (orange). The extended budget (green) writes the end of the
longest value, the physician's details, and all eleven values are exact. In the speculative mode
the verifier accepts all eleven unchanged.}
\label{fig:qualitative}
\end{figure*}

\Cref{fig:qualitative} traces one document: most of every value is committed by the first pass,
and the extended budget finishes the longest value.

\paragraph{With a verifier.}
In the speculative mode the base model verifies the drafted values, so each value follows the base
model's greedy prediction for that value. NAR-SD is 0.4 points below autoregressive decoding on average and 8.3$\times$ faster
(up to 10.4$\times$). It is also 3.3$\times$ faster than Fast-dVLM 3B, the other draft-and-verify
method, and 13.0 points more accurate, and both faster (3.2$\times$) and more accurate (+3.0) than
parallel-key AR. It is the only method in \cref{tab:main} that is both faster and more accurate than
the strongest autoregressive baseline.

% GENERATED by scripts/make_tables.py -- do not edit by hand.
\begin{table}[t]
\centering
\setlength{\tabcolsep}{3pt}
\resizebox{\linewidth}{!}{%
\begin{tabular}{lrrrrrr}
\toprule
 & \multicolumn{2}{c}{VAREX} & \multicolumn{2}{c}{CORD} & \multicolumn{2}{c}{FUNSD} \\
Value decoder & EM & ms & F1 & ms & F1 & ms \\
\midrule
CTC (ours) & \textbf{88.4} & \textbf{278} & \textbf{76.8} & \textbf{397} & \textbf{54.3} & \textbf{249} \\
Fixed-slot CE & 68.4 & 141 & 68.0 & 253 & 45.2 & 104 \\
\midrule
\multicolumn{7}{l}{Masked diffusion} \\
\quad 1 step & 63.3 & 150 & 65.9 & 256 & 41.0 & 146 \\
\quad 4 steps & 73.4 & 422 & 66.6 & 473 & 45.8 & 515 \\
\quad 8 steps & 79.9 & 793 & 69.6 & 779 & 52.8 & 1,000 \\
\quad 20 steps & 84.9 & 1,906 & 72.6 & 1,636 & 55.2 & 2,530 \\
\bottomrule
\end{tabular}}
\caption{\textbf{Value-decoder ablation.} Same base model, training data, epochs and inputs; only the value decoder differs. Every row uses the schema skeleton with predicted list counts, so the count pass is in every latency. Fixed-slot CE decodes each slot independently in one pass; masked diffusion unmasks the slots over the given number of steps, in blocks. Latency: ms per document.}
\label{tab:decoder}
\end{table}
  % generated; declared before 4.3 so the ablation tables float onto 4.3/4.4
% GENERATED by scripts/make_tables.py -- do not edit by hand.
\begin{table*}[!t]
\centering
\small
\setlength{\tabcolsep}{5pt}
\begin{tabular}{lrrrrrr}
\toprule
\multicolumn{7}{l}{\textit{(a) Ablations, NAR}} \\
 & \multicolumn{2}{c}{VAREX} & \multicolumn{2}{c}{CORD} & \multicolumn{2}{c}{FUNSD} \\
Variant & EM & ms & F1 & ms & F1 & ms \\
\midrule
Ours (stale prefix after pass 1) & \textbf{88.7} & \textbf{242} & \textbf{77.2} & \textbf{233} & \textbf{54.3} & \textbf{244} \\
\midrule
Full bidirectional, every pass & 88.6 & 394 & 77.2 & 351 & 54.5 & 400 \\
Stale prefix after pass 2 & 88.6 & 290 & 77.1 & 263 & 53.7 & 273 \\
Prefix-causal, trained & 86.5 & 271 & 73.8 & 256 & 54.8 & 266 \\
No confidence reveals ($T{=}1$) & 70.1 & 183 & 66.5 & 182 & 47.5 & 186 \\
10 slots per block & 83.0 & 251 & 73.7 & 234 & 52.6 & 243 \\
Loss on the last pass only & 84.8 & 255 & 68.9 & 248 & 50.0 & 254 \\
\midrule
\multicolumn{7}{l}{\textit{(b) Speculative mode, verification rounds}} \\
Rounds & EM & ms & F1 & ms & F1 & ms \\
\midrule
2 & 89.3 & 531 & 78.8 & 598 & 57.4 & 471 \\
\textbf{3 (ours)} & \textbf{91.0} & \textbf{646} & \textbf{80.4} & \textbf{686} & \textbf{59.0} & \textbf{582} \\
4 & 92.2 & 728 & 80.2 & 722 & 59.2 & 667 \\
to completion & 93.1 & 964 & 80.7 & 749 & 59.5 & 1,653 \\
\bottomrule
\end{tabular}
\caption{\textbf{Ablations and inference design.} (a) One change at a time against the paper model, with oracle list counts so that only the change differs; each model is evaluated with its own training settings. (b) Speculative mode by the cap on verification rounds, predicted list counts, drafts with $T{=}2$; values still open at the cap fall back to their first draft. Three rounds is our working point (\cref{tab:main}); \emph{to completion} has no cap and no fallback.}
\label{tab:design}
\end{table*}

% GENERATED by scripts/make_tables.py -- do not edit by hand.
\begin{table}[!t]
\centering
\setlength{\tabcolsep}{6pt}
\resizebox{\linewidth}{!}{%
\begin{tabular}{lrrr}
\toprule
Count pass & VAREX & CORD & FUNSD \\
\midrule
Count adapter & \textbf{88.4} & \textbf{76.8} & \textbf{54.3} \\
Oracle counts & 88.7 & 77.2 & 54.3 \\
Dual task & 88.2 & 74.7 & 54.2 \\
\bottomrule
\end{tabular}}
\caption{\textbf{Count pass.} Separate count adapter, oracle counts, and count queries trained into the extractor (dual task). NAR; EM on VAREX, F1 on the others.}
\label{tab:count}
\end{table}

% Decoding trace on VAREX test document cdph265 (CDPH 265), Tab. 1 recipes (2026-11-10: NAR extends
% only open values; SD drafts with T=2, three rounds). Colours are logged by the decode itself
% (TRACE_PROVENANCE in evaluate.run_nar_batch; scripts/trace_figure.py run -> scripts/trace_draw.py
% spans/draw --panels nar): which forward wrote each token -- confidence pass 1, passes 2-3, or the
% token-budget extension (window forward). The SD run of the same document changes no value, so
% only the NAR panel is drawn. Alternatives (GSA_3566, cdph266): fig/trace_*.pdf, preview_fig3_*.png.
  % decoding trace (scripts/trace_figure.py, trace_draw.py)
\subsection{Sequential cost vs document size}
\label{sec:steps}
% Forwards and latency vs document size. Data + plot: scripts/fwd_vs_keys.py (collect, plot).
\begin{figure}[t]
\centering
\includegraphics[width=\linewidth]{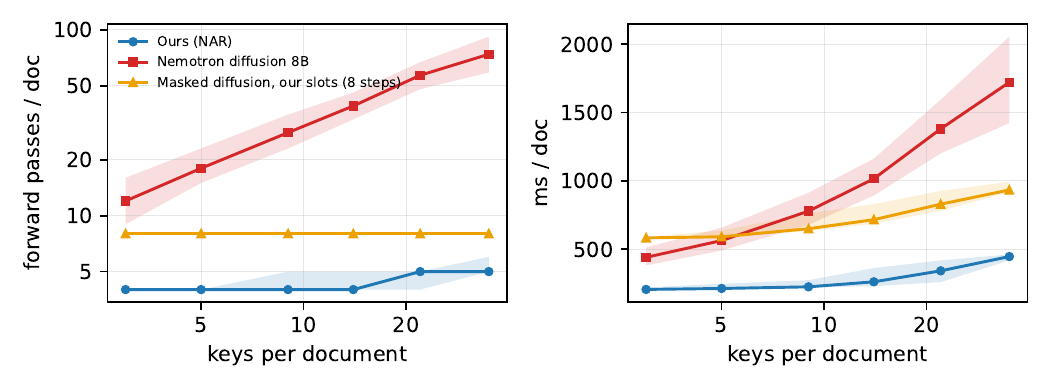}
\caption{\textbf{Sequential cost vs.\ document size.} Forward passes (left, log scale) and
latency (right) per document against the number of keys in its schema, pooled over VAREX,
CORD and FUNSD (1,924 documents; median and interquartile range per bin).
Our forward count stays at 4--6 for every document, while block diffusion's grows with the
size of the JSON it generates. Masked diffusion on our own skeleton (value-decoder ablation, 8
steps) is also flat but needs more forwards than CTC and is slower.}
\label{fig:fwd_vs_keys}
\end{figure}

Our number of forward passes should not grow with the size of the document (\cref{sec:parallel}).
We measure forwards and latency per document against the number of keys in its schema
(\cref{fig:fwd_vs_keys}). Across all 1,924 test documents, our extractor runs 4--6 forward passes per document regardless of how many keys
the schema has: the count pass, the confidence passes and up to two continuations. Block diffusion generates the answer JSON block by block, so its forward count follows
the length of what it writes: the median Nemotron document needs 12 forwards at 3 keys,
up to 74 at 31--50 keys, against 4--5 for ours. Latency grows for both models,
since our passes run over a longer skeleton when a schema has more keys, but more slowly for
ours: the median document takes 206--446 ms against 441--1,721 ms for Nemotron, so the gap widens with document size. 
% Our latency grows only because a longer schema makes each forward longer, not because more forwards are needed. 
The third curve is the masked-diffusion decoder of our
ablation (\cref{sec:fixed_slot}), run on our own schema skeleton. Its forward count does not grow
with the number of keys either, which confirms that the saving comes from decoding the values on a
given skeleton rather than generating the JSON in order. But it needs more forwards than CTC for
less accuracy: eight unmasking steps per block, plus extra blocks for long values, give 8.2--13.1
forwards per document on average across the benchmarks, against 4.5--5.2 for CTC, at 2.0--4.0$\times$
its latency and 8.5 points lower VAREX EM (\cref{tab:decoder}).

\subsection{Ablations}
\label{sec:ablations}

\paragraph{Value decoder.}
\label{sec:fixed_slot}
\Cref{tab:decoder} is a controlled comparison: base model, training data, epochs and inputs are the
same, and every row uses the predicted list counts. Fixed-slot decoding with a per-position
cross-entropy, the formulation of PIP~\cite{wang2026pip}, must place every token at its exact slot: per-slot cross-entropy fixes which slot each token
occupies, so every slot has to know how many tokens of the value precede it, whereas CTC scores
only the collapsed text and lets each token land in any slot.
It is faster than CTC, as it decodes in one pass, but trails by 20.0 points on VAREX, the benchmark with the most long and variably sized values,
and by 8.8--9.1 on CORD and FUNSD. Masked diffusion on the same slots~\cite{sahoo2024mdlm,nie2025llada} never matches CTC at
its speed: with four steps, already 1.2--2.1$\times$ slower, it trails by 8.5--15.0 points, and with
20 steps it is 4.1--10.2$\times$ slower and still behind on VAREX and CORD.

\paragraph{Model and inference design.}
\Cref{tab:design}a changes one thing at a time. Freezing the image--prompt key/values after the
first pass (\cref{sec:inference}) costs no accuracy: the stale prefix is within 0.2 points of full
bidirectional decoding and 1.5--1.6$\times$ faster. The image must attend to the skeleton, though:
training with a prefix-causal mask costs 2.2 EM on VAREX and 3.4 F1 on CORD, since the image is then encoded without
seeing which values the schema asks for. The confidence reveals are essential
(a single pass is 22--24\% faster but loses 6.8--18.6 points), and so are the 20-slot block (10 slots
lose 1.7--5.7) and the CTC loss on every pass (the last pass only loses 3.9--8.3, at the same
speed). The speculative mode keeps improving with more verification rounds, at the cost of latency
(\cref{tab:design}b). Without a cap it matches autoregressive decoding (it is 0.4--1.0 points more accurate on every
benchmark), but not its exact output: the skeleton always holds every key of the schema, while free decoding
sometimes omits one; the list counts come from the count pass; and each value is verified in the
context of the other values' drafts (\cref{sec:sd}). The two outputs differ on 8--26\% of the
fields, and the differences slightly favour the speculative mode (supplementary). It also costs up to
2.8$\times$ the latency of three rounds, since a few long values need many rounds. We use three
rounds as the working point: they keep most of that accuracy at the speed of \cref{tab:main};
values still open at the cap fall back to their draft.

\paragraph{Count pass.}
Predicting the list counts costs little (\cref{tab:count}): against oracle counts it loses 0.3 EM
on VAREX and 0.4 F1 on CORD, with 84\% of lists exact; its cost is 163 ms per CORD document.
Training the count queries into the extractor instead (one more epoch on value and count queries
together) removes the adapter and makes CORD 21\% faster, but
costs 2.1 F1 there (77\% of lists exact), so we keep the separate adapter. Count errors are mostly
under-counts and concentrate in nested lists (supplementary); a missed item costs only that item's
values, since the remaining items are still decoded.

\FloatBarrier   % flush all results floats before the conclusion (and the references)
\section{Conclusion}
\label{sec:conclusion}
We present a non-autoregressive extractor for schema-driven key information extraction that
decodes every value of a document at once. We exploit the fact that in key information extraction
the schema already gives a known frame for the answer, so regenerating it is redundant, only the
values have to be predicted. The frame is written out from the schema, each value gets its own
block of CTC slots so that its length is left to the model, and all values are refined together in
a fixed number of passes that does not grow with the document. The same extractor runs in two
modes, with no retraining: on its own, it is both more accurate and faster than Nemotron diffusion
8B at half its size; as a drafter for speculative decoding, verified by its own autoregressive
base, it beats Fast-dVLM on both counts and is the only method we evaluate that is faster and more
accurate than parallel-key autoregressive decoding.

No open model exists for non-autoregressive, schema-driven extraction, so we compared against the
closest general-purpose diffusion VLMs and against our own value-decoder variants trained on the
same data; the parallel decoder held up in both comparisons. We release the model and code as the
first open baseline for this task.

% CVPR 2027 mandatory LLM Usage Statement: after the main text, before the references; not counted
% toward the page limit.
\section*{LLM Usage Statement}
The research idea, the method and its design choices, the choice of experiments and ablations, and
the interpretation of the results are the authors' own; the authors directed the work at every step.
We used Claude (Anthropic), through the Claude Code agent, as an implementation and writing assistant
under that direction: it wrote and debugged training, decoding and evaluation code, set up and
monitored experiments on our cluster, produced the figures and tables with scripts that read the
logged evaluation outputs, and drafted and edited text, including grammar, which the authors revised.
The authors reviewed all results and the final text and take full responsibility for the paper.

{
    \small
    \bibliographystyle{ieeenat_fullname}
    \bibliography{refs}
}


\end{document}